%% file: acl2020.tex
\documentclass[11pt,a4paper]{article}

\usepackage[T1]{fontenc}
\usepackage[utf8]{inputenc}
\usepackage{authblk}
\usepackage[hyperref]{acl2020}
\usepackage{times}
\usepackage{latexsym}

\usepackage{booktabs}
\usepackage{url}
\usepackage{textcomp}
\usepackage{graphicx}
\usepackage{enumitem}
\usepackage{array,multirow,graphicx}
\setlist[itemize]{leftmargin=*}
\setlist[enumerate]{leftmargin=*}
\usepackage{microtype}
\usepackage{amsmath}

\newcommand{\findings}{\textsc{Findings }} \newcommand{\findingsD}{\textsc{Findings}} 
\newcommand{\rouge}{\textsc{Rouge }}

\newcommand{\impression}{\textsc{Impression }} 
\newcommand{\impressionD}{\textsc{Impression}} 
\newcommand{\oneS}{\ensuremath{{}^{\textstyle *}} }
\DeclareMathOperator{\softmax}{Softmax}

\aclfinalcopy

\title{Attend to Medical Ontologies: Content Selection for \\Clinical Abstractive Summarization}

\author[1]{\textbf{Sajad Sotudeh}}
\author[1]{\textbf{Nazli Goharian}}
\author[2]{\textbf{Ross W. Filice}}
\affil[1]{IR Lab, Georgetown University, Washington DC 20057, USA}
\affil[ ]{\normalsize \texttt {\{sajad, nazli\}@ir.cs.georgetown.edu}}

\affil[2]{MedStar Georgetown University Hospital, Washington DC 20007, USA} 
\affil[ ]{\normalsize \texttt{ross.w.filice@medstar.net}}
\usepackage{blindtext}

\newdimen\origiwspc
\newdimen\origiwstr

\date{}

\begin{document}

\maketitle
\begin{abstract}
Sequence-to-sequence (seq2seq) network is a well-established model for text summarization task. It can learn to produce readable content; however, it falls short in effectively identifying key regions of the source. In this paper, we approach the content selection problem for clinical abstractive summarization by augmenting salient ontological terms into the summarizer. Our experiments on two publicly available clinical data sets (107,372 reports of MIMIC-CXR, and 3,366 reports of OpenI) show that our model statistically significantly boosts state-of-the-art results in terms of \rouge metrics (with improvements: 2.9\% RG-1, 2.5\% RG-2, 1.9\% RG-L), in the healthcare domain where any range of improvement impacts patients' welfare.
\end{abstract}

\input{1-intro.tex}
\input{2-related}
\input{3-methods}
\input{4-exp}
\input{5-results}

\input{6-conc}
\input{7-ack}

% \aclfinalcopy % Uncomment this line for the final submission
\def\aclpaperid{***} %  Enter the acl Paper ID here

% \setlength\titlebox{5cm}
% You can expand the titlebox if you need extra space
% to show all the authors. Please do not make the titlebox
% smaller than 5cm (the original size); we will check this
% in the camera-ready version and ask you to change it back.

\bibliography{acl2020}
\bibliographystyle{acl_natbib}

\end{document}

%% file: 1-intro.tex
%     \begin{figure}
% \vspace{5pt}
% \centering \tiny
% \begin{tabular}{|p{7cm}|}\hline
% \vspace{1pt}
% \textbf{FINDINGS:} 
% LIVER: Liver is echogenic with slightly coarsened echotexture and mildly nodular contour. No focal lesion. Right hepatic lobe measures 14 cm in length.\\
% BILE DUCTS: No biliary ductal dilatation. Common bile duct measures 0.06 cm.\\
% GALLBLADDER: Partially visualized gallbladder shows multiple gallstones without pericholecystic fluid or wall thickening. ~~~~
% % VASCULATURE: TIPS:\\
% Proximal TIPS: 108 cm/sec, previously 82 cm/sec;
% Mid TIPS: 123 cm/sec, previously 118 cm/sec; 
% Distal TIPS: 85 cm/sec, previously 86 cm/sec; 
% PORTAL VENOUS SYSTEM:  [...]\\
% % Left portal vein: Patent with flow towards the TIPS; 48 cm/sec, previously 11 cm/sec.\\
% % Right portal vein: Not visualized.\\
% % IVC and hepatic venous system: Normal directional flow and waveform. Again, right hepatic vein is not visualized.\\
% \hline
% \vspace{1pt}
% \textbf{IMPRESSION: (Summary)}
% 1. Stable examination. Patent TIPS\\
% 2. Limited evaluation of gallbladder shows cholelithiasis.\\
% 3. Cirrhotic liver morphology without biliary ductal dilatation.\\
% \hline
% \end{tabular}
% \caption{\small{Abbreviated example of radiology note and its summary.}}
% \label{fig:note}
% \vspace{-12pt}
% \end{figure}

\section{Introduction}
    Radiology reports convey the detailed observations along with the significant findings about a medical encounter. Each radiology report contains two important sections:\footnote{Depending on institution, radiology reports may or may not include other fields such as \textsc{Background}.} \findings that encompasses radiologist's detailed observations and interpretation of imaging study, and \impression summarizing the most critical findings. \impression (usually couple of lines and thrice smaller than finding) is considered as the most integral part of report \cite{Ware2017EffectiveRR} as it plays a key role in communicating critical findings to referring clinicians. Previous studies have reported that clinicians mostly read the \impression as they have less time to review findings, particularly those that are lengthy or intricate \cite{Flanders2012RadiologyRA, Xie2019IntroducingIE}.

    In clinical setting, generating \impression from \findings can be subject to errors \cite{gershanik2011critical, Brady2016ErrorAD}. This fact is especially crucial when it comes to healthcare domain where even the smallest improvement in generating \impression can improve patients' well-being.
    Automating the process of impression generation in radiology reporting would save clinicians' read time and decrease fatigue \cite{Flanders2012RadiologyRA, Kovacs2018BenefitsOI} as clinicians would only need to proofread summaries or make minor edits. 
    % Motivated by these reasons, we hypothesize to automate the process of generating \impression given a report of \findingsD.
    \par
    Previously, \citet{MacAvaney2019OntologyAwareCA} showed that augmenting the summarizer with entire ontology (i.e., clinical) terms within the \findings can improve the content selection and summary generation to some noticeable extent. Our findings, further, suggest that radiologists select \textit{significant} ontology terms, but not all such terms, to write the \impressionD. Following this paradigm, we hypothesize that selecting the most \textit{significant} clinical terms occurring in the \findings and then incorporating them into the summarization would improve the final \impression generation. We further examine if refining \findings word representations according to the identified clinical terms would result in improved \impression generation.
    % In this paper, we step further toward identifying the most \textit{significant} ontological terms specific to a given report, and incorporate them into the abstractive summarizer.

    Overall, the contributions of this work are twofold: (i) We propose a novel seq2seq-based model to incorporate the salient clinical terms into the summarizer (\textsection\ref{sec:summarization_model}). We pose copying likelihood of a word as an indicator of its saliency in terms of forming \impressionD, which can be learned via a sequence-tagger (\textsection\ref{sec:content_selector}); (ii) Our model statistically significantly improves over the competitive baselines on MIMIC-CXR publicly available clinical dataset. To evaluate the cross-organizational transferability, we further evaluate our model on another publicly available clinical dataset (OpenI) (\textsection\ref{sec:results}).

    % To summarize, our contributions are listed as follows:
    % \begin{itemize}
    
    %     \item To the best of our knowledge, we are the first in clinical domain to adopt a novel content selector built upon \texttt{BERT} contextual embeddings to pick up the most important medical terms within the source that are likely to be copied into target.
        
    %     \item Enhancing information filtering process of source (i.e., \findings) by designing a gated ontological layer that modifies the word representation of source terms based off of salient ontological signals identified by content selector.
        
    %     \item Detailed analysis of system-generated and human-written summaries by an expert radiologist on a large collection of radiology reports and providing insights into the qualities of our model.
    % \end{itemize}

% \begin{itemize} 

%     \item How information are gathered in clinical notes, in particular radiology reports. Structure of radiology reports (Impression, Findings, Background, etc. ). The importance of Findings and Impression section, and how the summaries could be used in healthcare domain. 
%     \item Challenges in writing actual impression section by radiologists, ==> automating summarization process of clinical reports is of interest. 
%     \item Narrow down to our own work, like in this study we want to [...]. Brining up the methodolgy in a high-level description. 
%     \item A bit giving description of results if they're significant enought (which I assume they are)
%     \item Listing contributions of this work.
%     \item DIFFERENTS BETWEEN THIS WORK WITH PRIOR WORKS.
% \end{itemize}

%% file: 2-related.tex
\section{Related Work}

Few prior studies have pointed out that although seq2seq models can effectively produce readable content, they perform  poorly at selecting salient content to include in the summary \cite{Gehrmann2018BottomUpAS, Lebanoff2019ScoringSS}. Many attempts have been made to tackle this problem \cite{Zhou2017SelectiveEF, Lin2018GlobalEF, Hsu2018AUM, Lebanoff2018AdaptingTN, You2019ImprovingAD}. 
% % \citet{Cohan2018ADA} hypothesized to use hierarchical attention to identify relevant sections in a document. 
% % \citet{} used sentence-level attention to tune word-level attention in a sense that words within less attended sentences have less chance of being produced.
For example, \citet{Zhou2017SelectiveEF} used sentence representations to filter secondary information of word representation. Our work is different in that we utilize ontology representations produced by an additional encoder to filter word representations. \citet{Gehrmann2018BottomUpAS} utilized a data-efficient content selector, by aligning source and target, to restrict the model's attention to likely-to-copy phrases. In contrast, we use the content selector to find domain knowledge alignment between source and target. Moreover, we do not focus on model attention here, but on rectifying word representations.
% % \citet{You2019ImprovingAD} proposed a focus-attention mechanism to aid document encoding, and a saliency-selection network to wisely manage information flow from encoder to decoder. 
% \citet{Lebanoff2019ScoringSS} proposed scoring singletons and pairs of sentences that are summary-worthy and then fusing them to form summary sentences.

Extracting clinical findings from clinical reports has been explored previously \cite{Hassanpour2016InformationEF, Nandhakumar2017ClinicallySI}. For summarizing radiology reports, \citet{Zhang2018LearningTS} recently used a separate RNN to encode a section of radiology report.\footnote{\textsc{Background} field. }
% to guide the decoder in decoding process 
 Subsequently, \citet{MacAvaney2019OntologyAwareCA} extracted clinical ontologies within the \textsc{Findings} to help the model learn these useful signals by guiding decoder in generation process. Our work differs in that we hypothesize that all of the ontological terms in the \findings are not equally important, but there is a notion of \textit{odds of saliency} for each of these terms; thus, we focus on refining the \findings representations.
%  Unlike \citet{MacAvaney2019OntologyAwareCA} that guided the decoder in generation process, we focus on refining \findings word representations.
% We further employ a method to identify significant ontological terms specific to a report and then incorporate them into our abstractive summarizer.
% We then utilize a separate RNN to encode these ontological terms, yielding an ontology dense vector that is further used to enhace word representations of \findings section.

%% file: 3-methods.tex
% \section{Problem Specification}
% For radiology report summarization, given a passage of \findings $\mathbf{x} = \{x_1, x_2, ..., x_n\}$ where $n$ is the  length of \findingsD, our goal is to find an \impression sequence $\mathbf{y} = \{y_1, y_2, ..., y_m\}$ that summarizes significant clinical findings seen in \textrm{x}, with $m$ being the length of \impression and $m \ll n$. Here, the problem is to learn a mapping function $f(x)$ parameterized by $\theta$ that can maximize the probability of producing correct impressions. In this paper, we approach learning function $f(x)$ by a seq2seq neural model.
% \footnote{\impression is the technical word used in clinical domain, but we use }

\section{Model}

Our model consists of two main components: (1) a content selector to identify the most salient ontological concepts specific to a given report, and (2) a summarization model that incorporates the identified ontology terms within the \findings into the summarizer. The summarizer refines the \findings word representation based on salient ontology word representation encoded by a separate encoder.

% We first train a bi-LSTM built upon BERT embeddings to identify the most salient clinical concepts specific to a given report. We cast the content selection problem as a sequence-labeling task with the aim of identifying ontological terms within \findings that are likely to be copied into \impressionD. Then we add a separate ontology encoder to the standard seq2seq framework to encode identified ontological terms to a dense vector. This vector is further used to enhance the representations yielded by the findings encoder and produce ontology-aware representations. Finally, an attentional decoder decodes the \impression by attending over the ontology-aware representations. \par

% In this section, we elaborate on our model consisting of a content selector, and a summarization system to incorporate salient ontological terms identified by the content selector.

\subsection{Content Selector}
\label{sec:content_selector}
The content selection problem can be framed as a word-level extraction task in which the aim is to identify the words within the \findings that are likely to be copied into the \impressionD. We tackle this problem through a sequence-labeling approach. We align \findings and \impression to obtain required data for sequence-labeling task. To this end, let $b_1, b_2, ..., b_n$ be the binary tags over the \findings terms $\mathbf{x}=\{x_1, x_2, ..., x_n\}$, with $n$ being the length of the \findingsD. We tag word $x_i$ with 1 if it meets two criteria simultaneously: (1) it is an ontology term, (2) it is directly copied into \impressionD, and 0 otherwise. 
% Intuitively, ontological terms that are labeled with 1 (i.e., copied) are contributive in forming \impression since they have been selected as salient content within \findings that should also appear in the \impressionD. Hence, 
At inference, we characterize the copying likelihood of each \findings term as a measure of its saliency.

Recent studies have shown that contextualized word embeddings can improve the sequence-labeling performance \cite{Devlin2019BERTPO, Peters2018DeepCW}. To utilize this improvement for the content selection, we train a bi-LSTM network on top of the BERT embeddings with a softmax activation function. The content selector is trained to maximize log-likelihood loss with the maximum likelihood estimation. At inference, the content selector calculates the selection probability of each token in the input sequence. Formally, let $\mathcal{O}$ be the set of ontological words which the content selector predicts to be copied into the \impressionD:
\begin{equation}
\label{eq:cs}
    \mathcal{O} = \{o_i | o_i \in F_{\mathcal{U}}(\mathbf{x}) \land p_{o_i} \geq \epsilon\}
\end{equation}
where $F_{\mathcal{U}}(\mathbf{x})$ is a mapping function that takes in \findings tokens and outputs word sequences from input tokens if they appear in the ontology (i.e., RadLex)~\footnote{RadLex version 3.10, \url{http://www.radlex.org/Files/radlex3.10.xlsx}}, and otherwise skips them. $p_{o_i}$ denotes the selection probability of ontology word $o_i$, and $\epsilon \in [0,1]$ is the copying threshold.  
% Let $\mathbf{o}=F_\mathscr{O}(\mathbf{x})$ where $F_\mathscr{O}$ be a mapping function, e.g., a simple mapping function that only outputs a word sequence if it appears in the ontology and otherwise skips it.

\subsection{Summarization Model}
\label{sec:summarization_model}

\subsubsection{Encoders}
\label{subsec:enc}
We exploit two separate encoders: (1) findings encoder that takes in the \findingsD, and (2) ontology encoder that maps significant ontological terms identified by the content selector to a fix vector known as ontology vector. The findings encoder is fed with the embeddings of \findings words, and generates word representations $\mathbf{h}$. Then, a separate encoder, called ontology encoder, is used to process the ontology terms identified by the content selector and produce associated representations $\mathbf{h^o}$.
\begin{equation}
  \begin{array}{l}
    \mathbf{h} = \textrm{Bi-LSTM}(\mathbf{x})\\
    \mathbf{h^o} = \textrm{LSTM}(\mathbf{\mathcal{O}})
  \end{array}
\end{equation}
where $\mathbf{x}$ is the \findings text, $\mathcal{O}$ is the set of ontology terms occurring in the \findings and identified by the content selector, $\mathbf{h^o}=\{h^{o}_1, h^{o}_2, ..., h^{o}_l\}$ is the 
word representations yielded from the ontology encoder. Note that $h^{o}_l$--called ontology vector-- is the last hidden state containing summarized information of significant ontologies in the \findingsD.

% It should be noted that $h_i$ combines the hidden states from forward and backward directions. 
% \begin{equation}
%     \label{eq:ont_enc}
    
% \end{equation}

% For the sake of simplicity, we denote $h^{o}_m$ as $h^{ont}$ to be the ontology vector that preserves summarized information of significant ontological terms occurred in the \findingsD.
\begin{figure}
\centering
\includegraphics[scale=0.365]{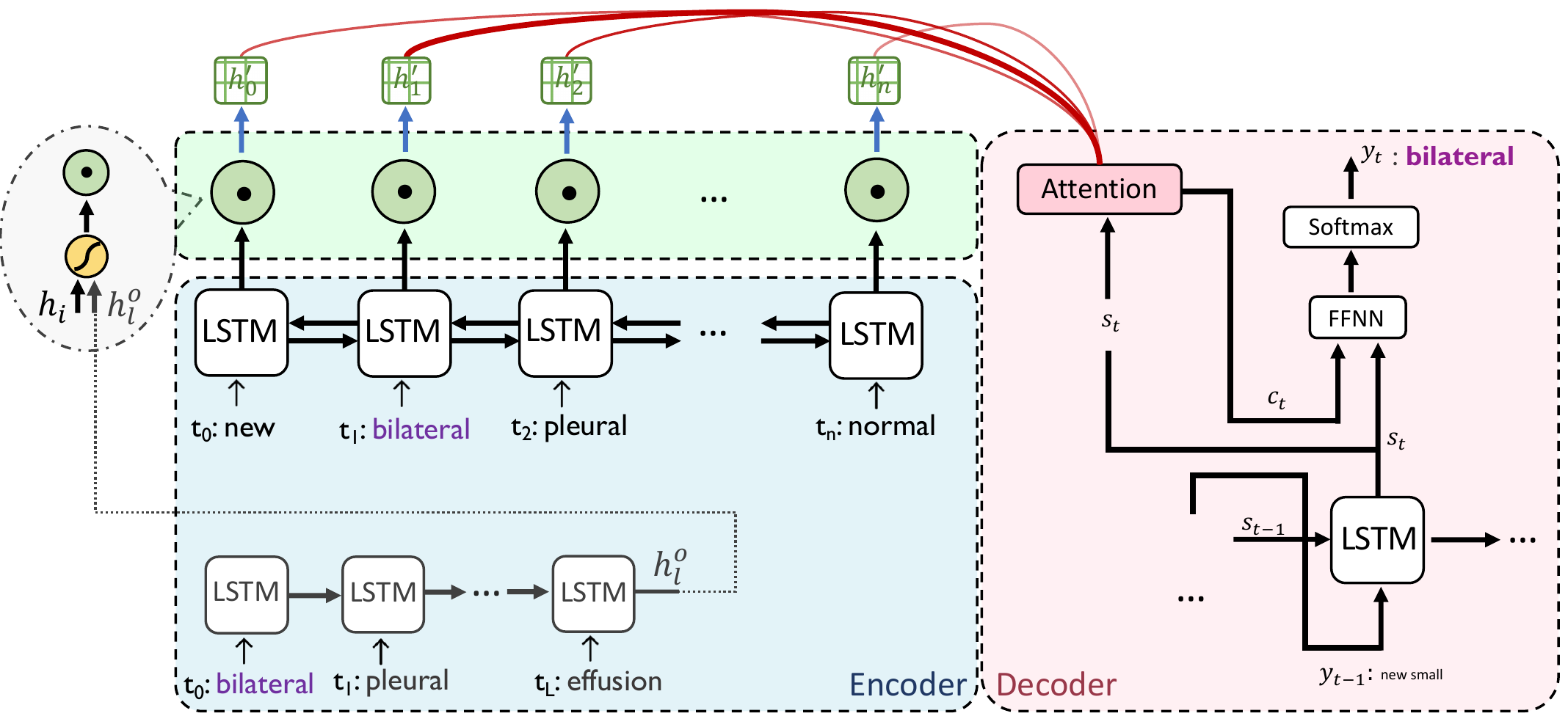}
\caption{ Overview of our summarization model. As shown, ``bilateral'' in the \findings is a significant ontological term which has been encoded into the ontology vector. After refining \findings word representation, the decoder computes attention weight (highest on ``bilateral'') and generates it in the \impressionD.
% After encoding \findingsD, the ontology vector $h^o_l$ is used along with \findings word hidden $h_i$ to develop a filtering gate on top of the findings encoder, which multiplies the \findings word hidden $h$.  Then the decoder generates the \impression by attending over ontology-aware hidden representations $h^\prime$.
}
\label{fig:summ_model}
\end{figure}

\subsubsection{Ontological Information Filtering}
Although de facto seq2seq frameworks implicitly model the information flow from encoder to decoder, the model should benefit from explicitly modeling the selection process. To this end, we implement a filtering gate on top of the findings encoder to refine the \findings word representations according to the significant ontology terms within the \findings and produce ontology-aware word representations. Specifically, the filtering gate receives two vectors: the word hidden representation $h_i$ that has the contextual information of word $x_i$, and the ontology vector $h^{o}_l$ including the overal information of significant ontology words within the \findingsD. The filtering gate processes these two vectors through a liner layer with Sigmoid activation function. We then compute the ontology-aware word hidden representation $h^\prime_i$, given the source word hidden representation $h_i$ and the associated filtering gate $F_i$.
\begin{equation}
    \begin{array}{l}
         F_i = \sigma(W_h[h_i;h^{o}_l] + b) \\
         h^{\prime}_i = h_i \odot F_i
    \end{array}
\end{equation}
where $W_h$ is the weight matrix, $b$ denotes the bias term, and
 $\odot$ denotes element-wise multiplication.

\subsubsection{Impression Decoder}
We use an LSTM network as our decoder to generate the \impression iteratively. In this sense, the decoder computes the current decoding state $\mathbf{s_t = LSTM(s_{t-1}, y_{t-1})}$, where $\mathbf{y_{t-1}}$ is the input to the decoder (human-written summary tokens at training, or previously generated tokens at inference) and $\mathbf{s_{t-1}}$ is the previous decoder state. The decoder also computes an attention distribution $\mathbf{a}=\softmax(\mathbf{\mathbf{{h^\prime}^\top \mathbf{V} s^\top}})$ with $\mathbf{h^\prime}$ being the ontology-aware word representations. The attention weights are then used to compute the context vector $\mathbf{c_t}=\sum_i^n a_i\mathbf{h^\prime_i}$ where $n$ is the length of the \findingsD. Finally, the context vector and decoder output are used to either generate the next token from the vocabulary or copy it from the \findingsD. 

%% file: 4-exp.tex
\section{Experiments}
% In this section, we introduce two real-world data collections we use for the experiments, followed by medical ontologies, baseline models, and parameter settings.

\subsection{Dataset and Ontologies}
\noindent \textbf{MIMIC-CXR. } This collection \cite{Johnson2019MIMICCXRAL} is a large publicly available dataset of radiology reports. Following similar report pre-processing as done in \cite{Zhang2018LearningTS}, we obtained 107,372 radiology reports.
% Concretely, we removed a report if: (1) either \findings or \impression is missing within the report; (2) \findings has less than 20 tokens, or \impression includes less than 3 tokens. 
For tokenization, we used ScispaCy \cite{Neumann2019ScispaCyFA}. We randomly split the dataset into 80\%(85,898)-10\%(10,737)-10\%(10,737) train-dev-test splits.\\
\noindent\textbf{OpenI.} A public dataset from the Indiana Network for Patient Care  \cite{DemnerFushman2016PreparingAC} with 3,366 reports. Due to small size, it is not suitable for training; we use it to evaluate the cross-organizational transferability of our model and baselines. \\
\noindent\textbf{Ontologies.} We use RadLex, a comprehensive radiology lexicon, developed by Radiological Society of North America (RSNA), including 68,534 radiological terms organized in hierarchical structure. 
% To capture ontological terms within reports, we use exact n-gram matching with the RadLex terms. 

\subsection{Baselines}
We compare our model against both known and state-of-the-art extractive and abstractive models.
\begin{itemize}[leftmargin=*,label={-}]
% \vspace{-0.42em}

\item \textbf{LSA}~\cite{Steinberger2004LSA}: An extractive vector-based model that employs Sigular Value Decomposition (SVD) concept.
% \vspace{-0.42em}

\item \textbf{NeuSum}~\cite{Zhou2018ND}: 
A state-of-the-art extractive model  that integrates 
the process of source sentence scoring and selection.\footnote{We use open code at \url{https://github.com/magic282/NeuSum} with default hyper-parameters.}
% \vspace{-0.4em}

\item \textbf{Pointer-Generator (PG)}~\cite{See2017GetTT}: An abstractive summarizer that extends ses2seq networks by adding a copy mechanism that allows for directly copying tokens from the source. 
% \vspace{-0.41em}

\item \textbf{Ontology-Aware~Pointer-Generator~(Ont. PG)}~\cite{MacAvaney2019OntologyAwareCA}: An extension of PG model that first encodes \emph{entire} ontological concepts within \findingsD, then uses the encoded vector to guide decoder in summary decoding process.
% \vspace{-0.41em}

\item \textbf{Bottom-Up Summarization (BUS)} ~\cite{Gehrmann2018BottomUpAS}: An abstractive model which makes use of a content selector to constrain the model's attention over source terms that have a good chance of being copied into the target.\footnote{We re-implemented the BUS model.}
\end{itemize}
% \vspace{-0.41em}

\subsection{Parameters and Training}
We use \textsc{SciBert} model  \cite{Beltagy2019SciBERTAP} which is pre-trained over biomedical text. We employ 2-layer bi-LSTM encoder with hidden size of 256 upon \textsc{Bert} model. 
The dropout is set to 0.2. We train the network to minimize cross entropy loss function, and optimize using Adam optimizer \cite{Diedrik2014Adam} with learning rate of $2e^{-5}$.

For the summarization model, we extended on the open base code by \citet{Zhang2018LearningTS} for implementation.\footnote{\url{https://github.com/yuhaozhang/summarize-radiology-findings}} We use 2-layer bi-LSTM, 1-layer LSTM as findings encoder, ontology encoder, and decoder with hidden sizes of 200 and 100, respectively. We also exploit 100d GloVe embeddings pretrained on a large collection of 4.5 million radiology reports \cite{Zhang2018LearningTS}. We train the network to optimize negative log likelihood with Adam optimizer and a learning rate of 0.001.
% We re-implement the BUS model based on \citet{Gehrmann2018BottomUpAS}, and to make it comparable with our model, we use the same hyper-parameters. 

%% file: 5-results.tex
\begin{table}[t]
\centering 
\begin{tabular}{llll}
\toprule
 Method                    & RG-1  & RG-2  & RG-L  \\
\midrule
% \multirow{2}{*}{\STAB{\rotatebox[origin=c]{90}{\scriptsize	 Extractive}}} 
% & LexRank \cite{Erkan2004LexRankGL}                 &  & 18.87 & 8.48 & 17.72 \\
 LSA                        & 22.21 & 11.17 & 20.80 \\
 \textsc{NeuSum}              & 23.97 & 12.82 & 22.61 \\
\midrule
% \multirow{4}{*}{\STAB{\rotatebox[origin=c]{90}{\scriptsize Abstractive}}} 

 PG                              & 51.20 & 39.13 & 50.16 \\
 Ont. PG                    & 51.84 & 39.59 & 50.72 \\
 BUS    & 52.04     & 39.69     & 50.83     \\
 Ours (this work)                      & \textbf{53.57\oneS} & \textbf{40.78\oneS} & \textbf{51.81\oneS} \\
\bottomrule
\end{tabular}
\caption{\rouge results on MIMIC-CXR. \oneS shows the statistical significance (paired t-test, $p<0.05$).}
\label{tab:final}
\end{table}

\section{Results and Discussion}
\label{sec:results}

\subsection{Experimental Results}
Table. \ref{tab:final} shows the \rouge scores of our model and baseline models on MIMIC-CXR, with human-written \textsc{Impressions} as the ground truth. Our model significantly outperforms all the baselines on all \rouge metrics with 2.9\%, 2.5\%, and 1.9\% improvements for RG-1, RG-2, and RG-L, respectively. While \textsc{NeuSum} outperforms the non-neural LSA in extractive setting, the extractive models lag behind the abstractive methods considerably, suggesting that human-written impressions are formed by abstractively selecting information from the findings, not merely extracting source sentences. When comparing Ont. PG with our model, it turns out that indeed our hypothesis is valid that a pre-step of identifying significant ontological terms can improve the summary generation substantially. As pointed out earlier, we define the saliency of an ontological term by its copying probability. 

As expected, BUS approach achieves the best results among the baseline models by constraining decoder's attention over odds-on-copied terms, but still underperforms our model. This may suggest that the intermediate stage of refining word representations based on the ontological word would lead to a better performance than superficially restricting attention over the salient terms. Table. \ref{tab:cs} shows the effect of content selector on the summarization model. For the setting without content selector, we encode all ontologies within the \findingsD. As shown, our model statistically significantly improves the results on RG-1 and RG-2.

% While Bottom-Up approach improves over PG by forcing the model to make decisive decisions to copy the words, our model polishes up the source word representations based of off ontological concepts --identified to be significant by the content selector-- and improves the model's both copy or generation capability.

\begin{table}[t!]
\centering
\begin{center}
\begin{tabular}{llll}
\toprule 
Method                   & RG-1  & RG-2  & RG-L  \\
\midrule
% PG                        &  & 39.50 & 20.29 & 39.27 \\
% Ont. PG               &  & - & - & - \\
BUS   & 40.02     & 21.89     & 39.37     \\
% \midrule
Ours (this work)                     & \textbf{40.88\oneS} & \textbf{24.44\oneS} & \textbf{40.37\oneS} \\
\bottomrule
\end{tabular}
\end{center}
\caption{\rouge results on Open-I dataset, comparing our model with the best-performing baseline. \oneS shows the statistical significance (paired t-test, $p<0.05$).}
\label{tab:openi}
\end{table}

\begin{table}[t!]
\centering
\begin{center}
\begin{tabular}{lllll}
\toprule
Setting                  &  & RG-1  & RG-2  & RG-L  \\
\midrule
% Ont. PG               &  & - & - & - \\
w/o Cont. Sel. &  & 52.47     & 40.11     & 51.39     \\
% \midrule
w/ Cont. Sel.                    &  & \textbf{53.57\oneS} & \textbf{40.78\oneS} & \textbf{51.81} \\
\bottomrule
\end{tabular}
\end{center}
\caption{\rouge results showing the impact of content selector in summarization model. \oneS shows the statistical significance (paired t-test, $p<0.05$).}
\label{tab:cs}
\end{table}

\begin{figure*}
\centering
\bgroup
\setlength{\tabcolsep}{0pt}
\renewcommand{\arraystretch}{0}
\begin{tabular}{c@{\hskip 0.5in}c@{\hskip 0.5in}c}
\includegraphics[scale=0.31]{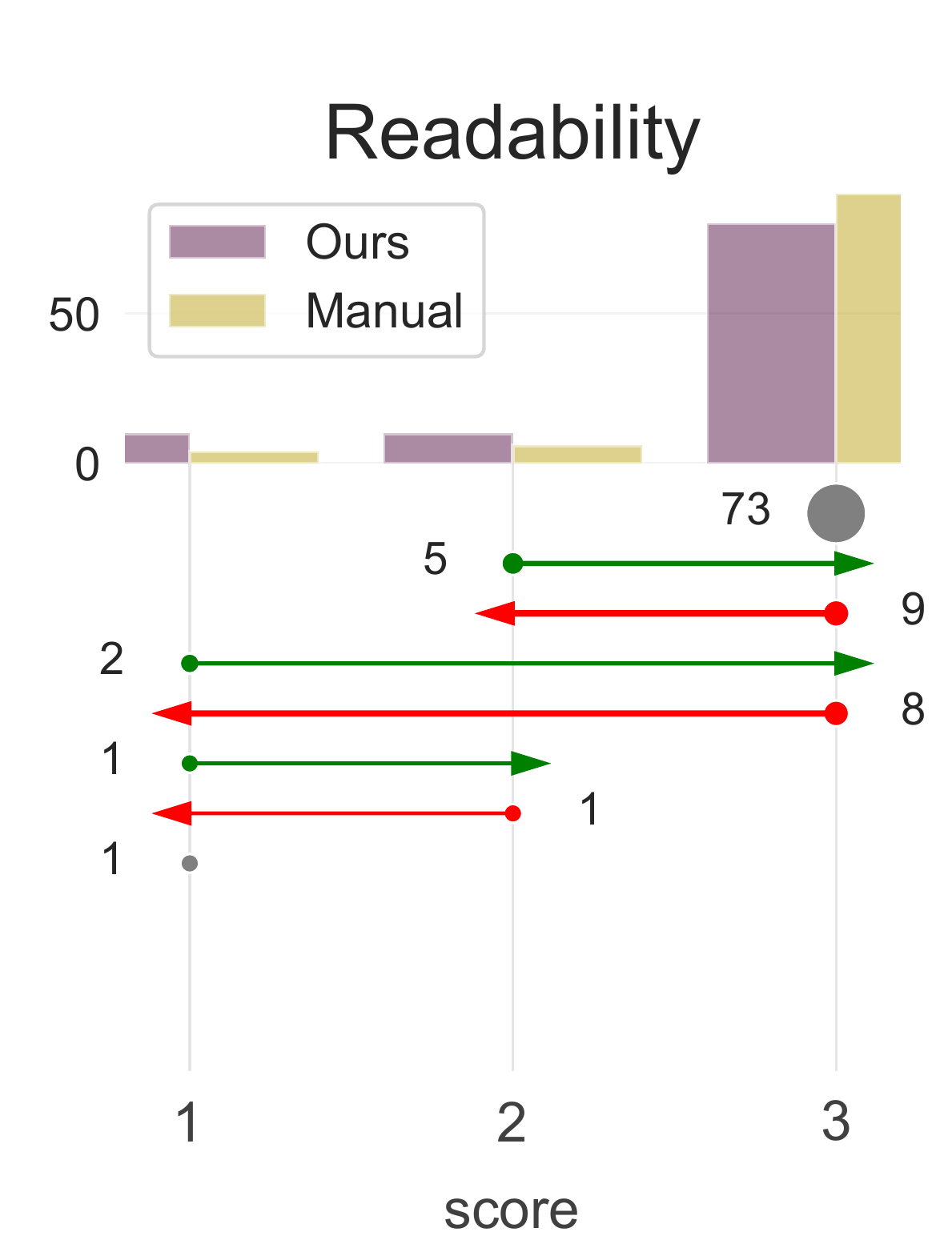} &
\includegraphics[scale=0.31]{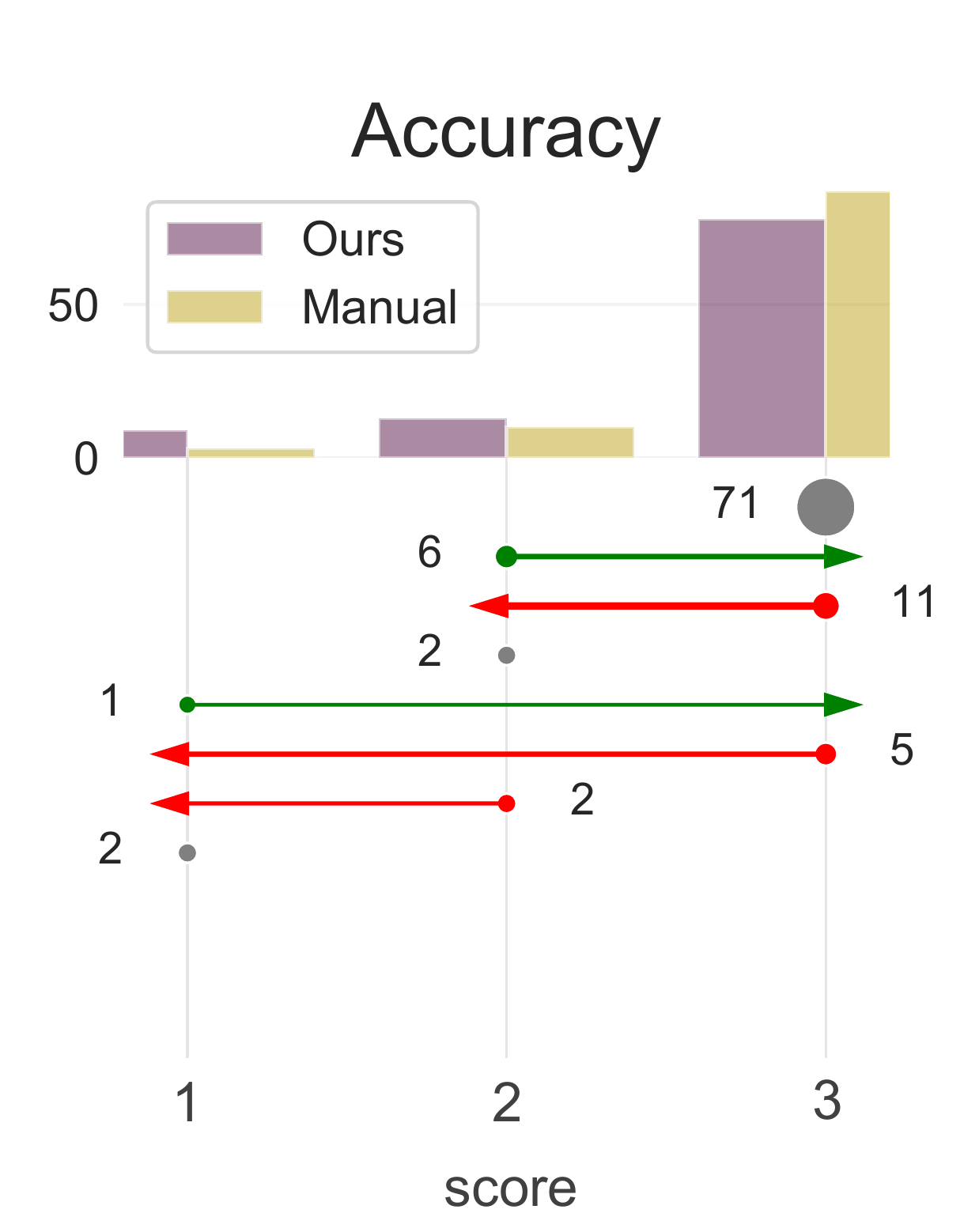} &
\includegraphics[scale=0.31]{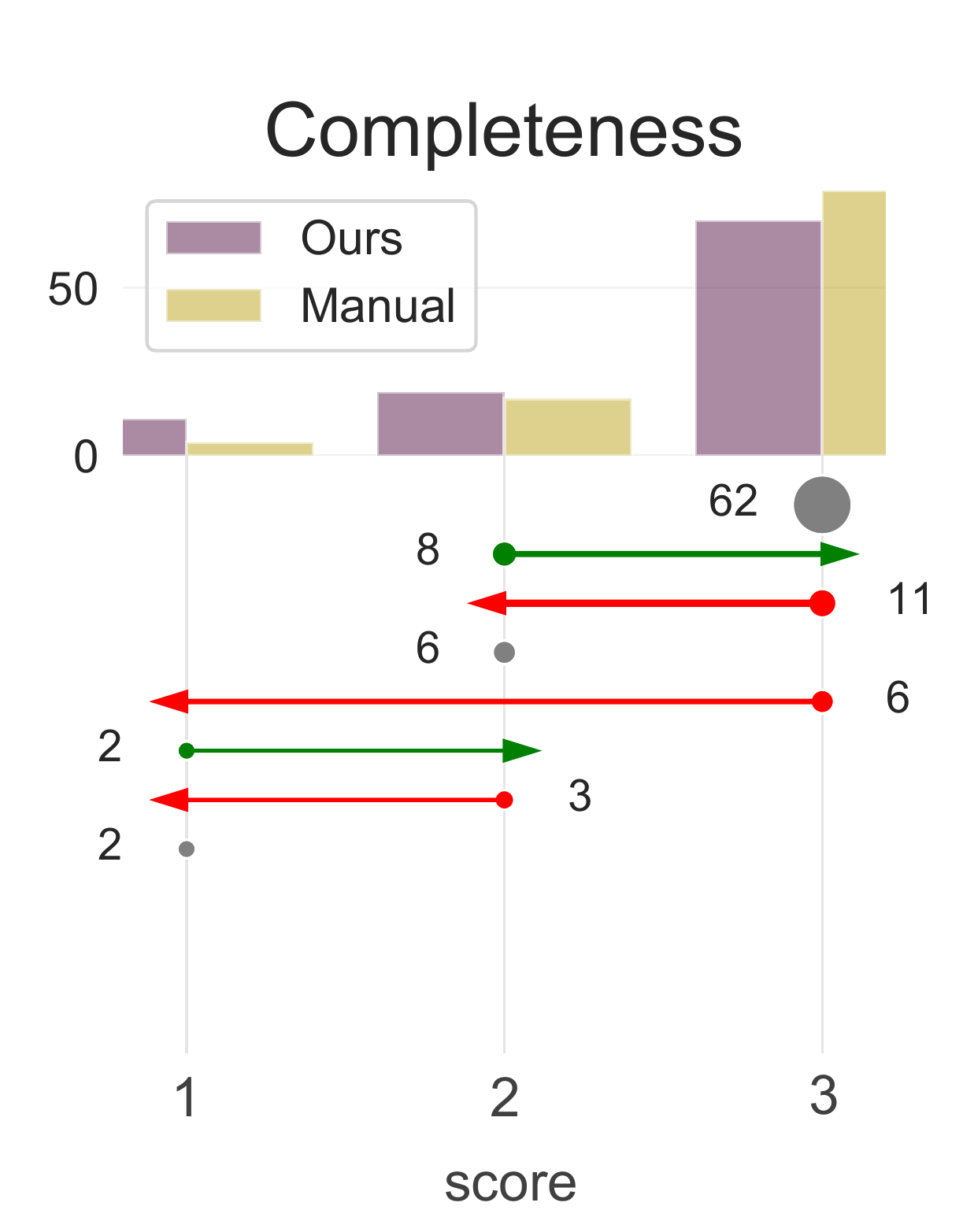} \\
(a) & (b) & (c) \\
\end{tabular}
\vspace{-0.5em}
\caption{Histograms and arrow plots showing differences between \impression of 100 manually-scored radiology reports. Although challenges remain to reach human parity for all metrics, 81\%~(a), 82\%~(b), and 80\%~(c) of our system-generated Impressions are as good as human-written Impressions across different metrics.}
\label{fig:arrows}
\egroup
\end{figure*}

% \begin{table}[h]
% \centering \small
% \begin{tabular}{llll}
% \toprule 
%                      & RG-1 & RG-2 & RG-L \\ \midrule
% w/o Content Selector &   47.57   &   35.80   &   46.73  \\ 
% w/ Content Selector  &   \textbf{53.91}   &   \textbf{40.67}   &   \textbf{52.91}   \\ \bottomrule

% \end{tabular}
% \caption{Effect of the content selector on the summarization model}
% \label{tab:cs}
% \end{table}

% Table. \ref{tab:cs} shows the evaluation performance of our summarization model in the absence and the presence of content selection component. For the setting without content selector, we simply encode all of the onotological terms within the \findingsD. It can be observed that our summarization model performs out-of-the-box when it comes along with the content selector. This suggests that encoding ontological words that are contibutive in forming the final \impression (likely to be copied into the \impression) is notably effective at predicting the correct \impression sequence.

% \subsection{Cross-Organizational Transferability}
% \label{subsec:transferability}
To further evaluate the transferability of our model across organizations, we perform an evaluation on OpenI with our best trained model on MIMIC-CXR. As shown in Table. \ref{tab:openi}, our model significantly outperforms the top-performing abstractive baseline model suggesting the promising cross-organizational transferability of our model.

\subsection{Expert Evaluation}
While our approach achieves the best \rouge scores, we recognize the limitation of this metric for summarization task~\cite{Cohan2016RevisitingSE}. To gain a better understanding of qualities of our model, we conducted an expert human evaluation. To this end, we randomly sampled 100 system-generated Impressions with their associated gold from 100 evenly-spaced bins (sorted by our system's RG-1) of MIMIC-CXR dataset. The Impressions were shuffled to prevent potential bias. We then asked three experts~\footnote{Two radiologists and one medical student.} to score the given Impressions independently on a scale of 1-3 (worst to best) for three metrics: \emph{Readability.} understandable or nonsense; \emph{Accuracy.} fully accurate, or containing critical errors; \emph{Completeness.} having all major information, or missing key points.

Figure.~\ref{fig:arrows} presents the human evaluation results using histograms and arrow plots as done in ~\cite{MacAvaney2019OntologyAwareCA}, comparing our system's Impressions versus human-written Impressions. The histograms indicate the distribution of scores, and arrows show how the scores changed between ours and human-written. The tail of each arrow shows the score of human-written \impression, and its head indicates the score of our system's \impressionD. The numbers next to the tails express the count of Impressions that gained score of $\mathbf{s^\prime}$ by ours and $\mathbf{s}$ by gold.~\footnote{$\mathbf{s},\mathbf{s^\prime} \in \{1,2,3\}$} We observed that while there is still a gap between the system-generated and human-written Impressions, \emph{over 80\%} of our system-generated Impressions are as good~\footnote{Either tied or improved.} as the associated human-written Impressions. Specifically, 73\% (readability), and 71\% (accuracy) of our system-generated Impressions ties with human-written Impressions, both achieving full-score of 3; nonetheless, this percentage is 62\% for completeness metric. The most likely explanation of this gap is that deciding which findings are more important (i.e., should be written into Impression) is either subjective, or highly correlates with the institutional training purposes. Hence, we recognize cross-organizational evaluations in terms of Impression completeness as a challenging task. We also evaluated the inter-rater agreement using Fleiss' Kappa~\cite{Fleiss1971MeasuringNS} for our system's scores and obtained 52\% for readability, 47\% for accuracy, and 50\% for completeness, all of which are characterized as moderate agreement rate.

%% file: 6-conc.tex
\section{Conclusion}

We proposed an approach to content selection for abstractive text summarization in clinical notes. We introduced our novel approach to augment standard summarization model with significant ontological terms within the source. Content selection problem is framed as a word-level sequence-tagging task. The intrinsic evaluations on two publicly available real-life clinical datasets show the efficacy of our model in terms of \rouge metrics. Furthermore, the extrinsic evaluation by domain experts further reveals the qualities of our system-generated summaries in comparison with gold summaries.

%% file: 7-ack.tex
\section*{Acknowledgement}
We thank Arman Cohan for his valuable comments on this work. We also thank additional domain expert evaluators: Phillip Hyuntae Kim, and Ish Talati.